\documentclass[11pt,a4paper]{article}
\pdfoutput=1
\usepackage{graphicx}
\usepackage[hyperref]{acl2019}
\usepackage{times}
\usepackage{latexsym}

\usepackage{tabularx}
\usepackage{booktabs}
\usepackage{lingmacros}
\usepackage{bbding}
\usepackage{pifont}
\usepackage{latexsym}
\usepackage{tabularx}
\usepackage{amssymb}
\usepackage{amsmath}
\usepackage{bm}
\usepackage{amsthm}
\usepackage{multirow}
\usepackage{algorithm}
\usepackage{caption}
\usepackage{algorithmic}

\usepackage{url}
\def\aclpaperid{42}

\aclfinalcopy

\title{Diamonds in the Rough: Generating Fluent Sentences \\ from Early-Stage Drafts for Academic Writing Assistance}

\author{Takumi Ito\thanks{$\quad$The authors contributed equally}$^{\;\;,1,2}$ , Tatsuki Kuribayashi\footnotemark[1]$^{\;\;,1,2}$, Hayato Kobayashi$^{3,4}$, \\ 
{\bf Ana Brassard$^{4,1}$, Masato Hagiwara$^{5}$, Jun Suzuki$^{1,4}$,} and {\bf Kentaro Inui$^{1,4}$}  \\
 $^1$Tohoku University 
 $^2$Langsmith Inc.
 $^3$Yahoo Japan Corporation 
 $^4$RIKEN
 $^5$Octanove Labs LLC \\
 \texttt{\{t-ito, kuribayashi, jun.suzuki, inui\}@ecei.tohoku.ac.jp } \\
 \texttt{hakobaya@yahoo-corp.jp}, 
 \texttt{ana.brassard@riken.jp} \\
 \texttt{masato@octanove.com} \\}

\date{}

\begin{document}
\maketitle
\begin{abstract}
The writing process consists of several stages such as drafting, revising, editing, and proofreading.
Studies on writing assistance, such as grammatical error correction (GEC), have mainly focused on sentence \textit{editing} and \textit{proofreading}, where surface-level issues such as typographical, spelling, or grammatical errors should be corrected.
We broaden this focus to include the earlier \textit{revising} stage, where sentences require adjustment to the information included or major rewriting and propose \textit{Sentence-level Revision (SentRev)} as a new writing assistance task.
Well-performing systems in this task can help inexperienced authors by producing fluent, complete sentences given their rough, incomplete drafts.
We build a new freely available crowdsourced evaluation dataset consisting of incomplete sentences authored by non-native writers paired with their final versions extracted from published academic papers for developing and evaluating SentRev models.
We also establish baseline performance on SentRev using our newly built evaluation dataset.
\end{abstract}

\section{Introduction}
\label{sec:intro}
Academic writing can be a daunting task, even for experienced writers with a native or near-native command of English. Inexperienced, non-native speakers find themselves in an even more difficult situation---in addition to grammatical or spelling errors, their sentences may lack fluidity, have an awkward style, contain collocation errors, or have missing words where they could not remember or did not know the appropriate expressions. Such authors, especially students with insufficient academic experience, may often have difficulty putting their ideas and findings into words, even if the ideas are sound and contribute to the research community. Improving writing quality is thus a concern for both individual researchers and the academic community. 

Writing assistance technologies have been extensively studied in the natural language processing (NLP) field \citep{Brill2000, ng2014, grangier2018}. 
We focus on helping inexperienced authors in writing fluent grammatical sentences.

\begin{figure}[t]
    \includegraphics[width=7.5cm]{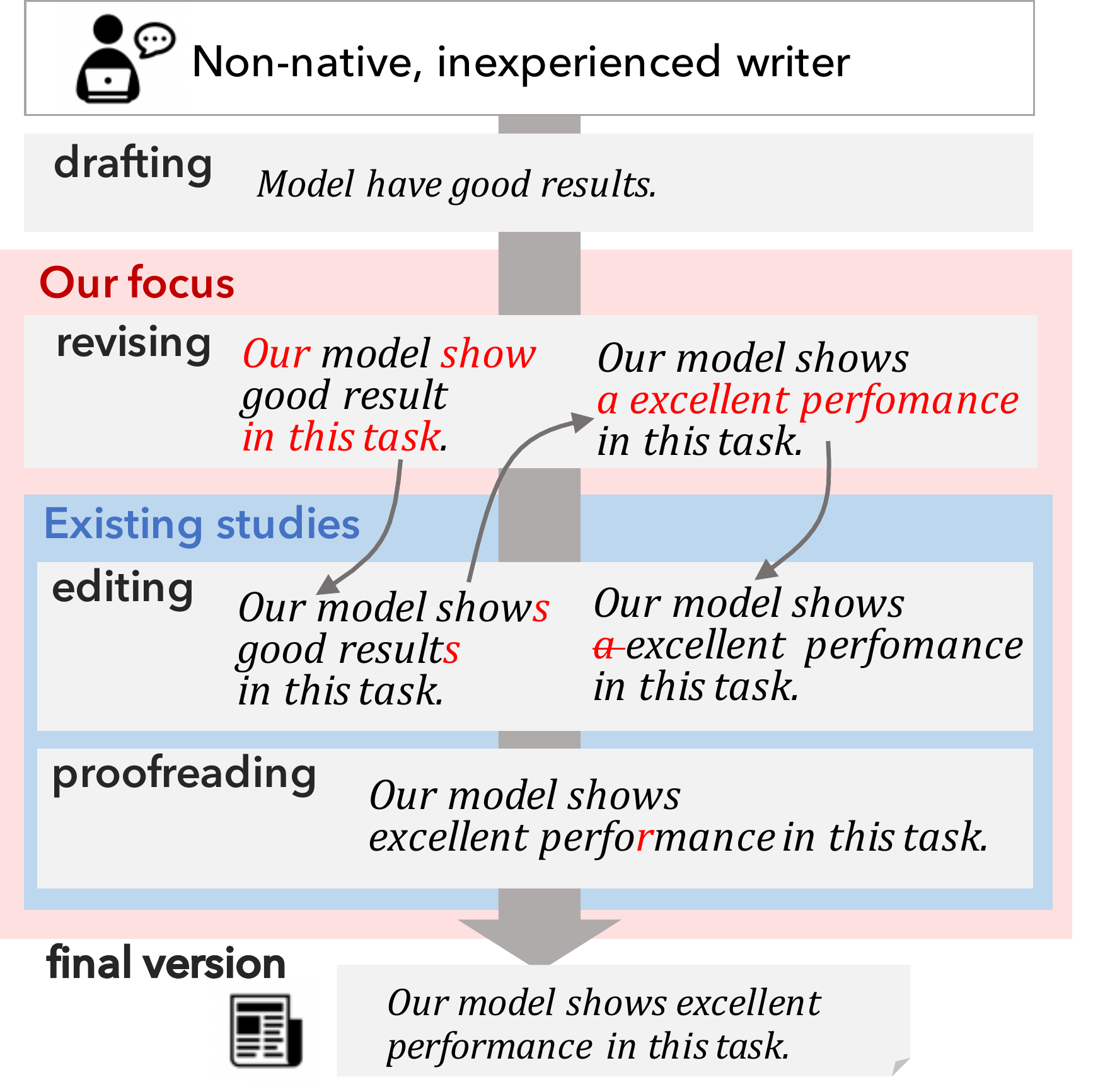}
    \centering
      \vspace{-0.1cm}
      \caption{Overview of the estimated process of writing a sentence {\it Our model shows excellent performance in this task.}. Writing activity consists of four stages: (i) drafting, (ii) revising, (iii) editing, and (iv) proofreading. }
      \label{fig:writing_process}
\end{figure}

Models developed for academic writing assistance using existing datasets can serve as a support system during the final stages by editing a nearly finished version of the draft. For example, \citet{daudaravicius2015aesw} collects scientific papers before and after professional editing from publishing companies, and \citet{dale2011helping} extract already published papers that still contain errors and correct the errors to obtain target fragments of text. 

Process-writing pedagogy, however, asserts that writing comprises several processes~\citep{susser1994process, seow2002writing, buchman2000power} as shown in Figure~\ref{fig:writing_process}.
This study takes on the challenge of automatic assistance in both the final checking process ({\it proofreading} and {\it editing}) and the earlier stages of writing ({\it revising}).
In the revising stage, authors may drastically modify the wording and supplement some words, a highly demanding task for non-native or less experienced writers. Assistance in this stage has been less explored in NLP.

In this study, we design a new type of academic writing assistance task, Sentence-level Revision (SentRev), where a system receives an early draft of a sentence, and generates a revised, error-free, proofread version. 

A critical issue in tackling this type of assistance task is that evaluation resources are scarce since early-stage draft sentences are not usually publicly available. 
To overcome this limitation, we release an evaluation dataset of pairs of draft sentences and their final versions, the~\textit{Set of Modified Incomplete TecHnical paper sentences} (\textsc{Smith}), that we created using crowdsourcing techniques.
Additionally, we evaluate the quality of our dataset and extensively analyze the characteristics of the obtained drafts. 
Finally, we train unsupervised models and report the baseline performance for our task on the \textsc{Smith} evaluation dataset.

Our contribution is fourfold:
\vspace{-0.2cm}
\begin{itemize}
 \setlength{\parskip}{0cm}
 \setlength{\itemsep}{0cm}
    \item We propose a new task---SentRev.
    \item We create an evaluation dataset, \textsc{Smith}, for SentRev using a new crowdsourcing approach and release it.\footnote{\url{https://github.com/taku-ito/INLG2019_SentRev}} 
    \item We compare the characteristics of our dataset with major corpora and analyze the obtained draft sentences.
    \item We establish baseline scores for SentRev.
\end{itemize}
\vspace{-0.2cm}

\section{The Sentence-level Revision task}
\label{sec:task}

\begin{table}[t]
    \centering
    \renewcommand{\arraystretch}{0.9}
    \small{
        \begin{tabularx}{7cm}{lX}
        \toprule
        Draft &
            {\it However, the F1 score of KBP 2017 coupus \textbf{\texttt{<*>}} decreased by the sub event base rule.} \\ 
        \cmidrule(lr){1-1}  \cmidrule(lr){2-2}
        Reference & 
            {\it However, subevent based constraints slightly reduced the F1 scores on KBP 2017 corpus.} \\ \midrule
        
        Draft & 
            {\it But, there are some important difference to \textbf{\texttt{<*>}} our work unique.} \\
        \cmidrule(lr){1-1}  \cmidrule(lr){2-2}
        Reference & 
            {\it However, there exist several key differences that make our work unique.} \\
        \bottomrule
        \end{tabularx}
    }
\caption{Examples of sentence-level revisions in our \textsc{Smith} dataset. Our task is to transform the draft sentences into their corresponding reference sentences.} 
\label{tab:task}
\end{table}

The proposed task, SentRev, is revising and editing incomplete draft sentences to create final versions.
Examples of sentence-level revision are shown in Table~\ref{tab:task}.

A draft sentence, $x$, may have several types of problems.
Surface-level problems such as typographical errors, spelling errors, or grammatical errors are a common occurrence. 
Wording problems, such as collocation errors or expressions being stylistically odd or inappropriate for the academic domain, are also typical of rough sentences written by non-native, inexperienced writers. 
The third type of error is \emph{information gaps}. 
Information gaps are cases where the author likely could not find the appropriate wording for the idea he or she wanted to convey, such as a specific expression common in the academic domain or a technical term. 
In addition, a draft sentence may be missing sections without the author being aware of this.
Solving the aforementioned problems in a draft sentence would elevate the draft sentence $x$ to its final or nearly final version $y$ with greatly improved correctness and fluency. 
Ideally, a single error-free and correctly filled-in final version should be generated while considering the context of the sentence. 
However, as a first step, an assistance system may output a set of \emph{likely candidates} for the user to choose from or be inspired by, which would be realistic for a real-world application. 

Our proposed task is, therefore, to generate likely final versions $y$ from early-draft sentences $x$.
For this purpose, we provide an evaluation dataset, \textsc{Smith}, comprising pairs of drafts and their final versions ($X$, $Y$).

\section{The \textsc{Smith} dataset}
\label{sec:data_construction_of_smis}

\begin{figure}[t]
    \centering
      \includegraphics[width=7.5cm]{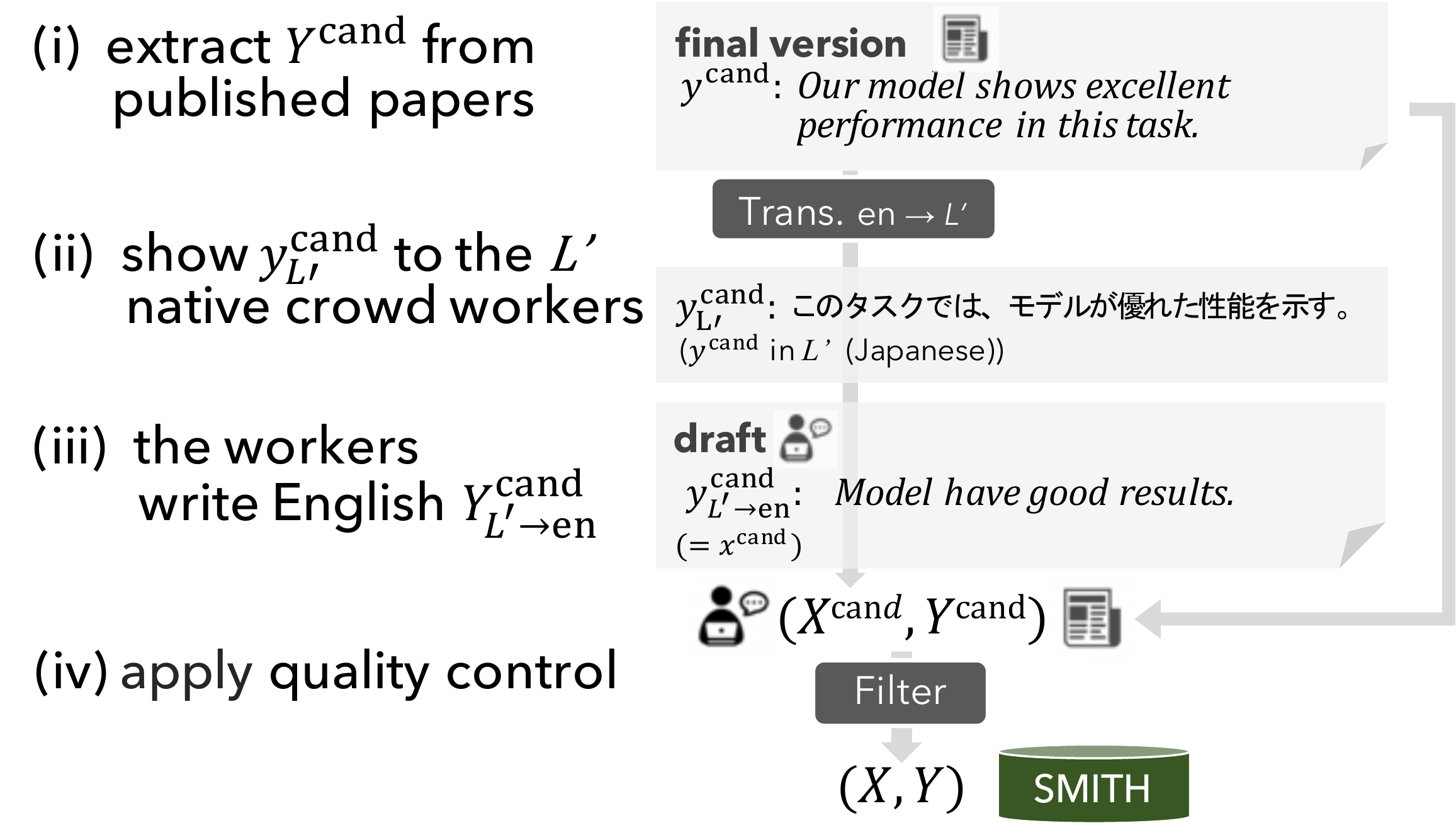}
      \vspace{-0.1cm}
      \caption{Overview of the crowdsourcing protocol for creating an evaluation dataset for the SentRev task.}
      \label{fig:crowd}
\end{figure}

\subsection{Dataset creation}
\paragraph{Process overview}
Although we cannot collect ``drafts" $X$ from published papers, we can easily collect the ``final versions" $Y$. 
We also have access to non-native, inexperienced writers through crowdsourcing services. 
Our test set creation process combines these two factors (Figure~\ref{fig:crowd}).
The protocol consists of the following four phases: 
\vspace{-0.2cm}
\begin{enumerate}
   \setlength{\parskip}{0cm} %
   \setlength{\itemsep}{0cm} %
    \item[(i)] Collecting a large number of sentences written by experts $Y^{\text{cand}}$ from published papers.
    \item[(ii)] Translating them into another language $L'$, resulting in sentences $Y^{\text{cand}}_{L'}$.
    \item[(iii)] Asking native speakers of $L'$ to translate $Y^{\text{cand}}_{L'}$ back into English $Y^{\text{cand}}_{L' \to \text{en}}$ through crowdsourcing. Henceforth, we denote $Y^{\text{cand}}_{L' \to \text{en}}$ as $X^{\text{cand}}$.
    \item[(iv)] Filtering the pairs of ($X^{\text{cand}}$, $Y^{\text{cand}}$) to ensure the quality of the dataset ($X$, $Y$).
\end{enumerate}

\vspace{-0.2cm}

This setting is analogous to the situation non-native writers face, as \citet{cohen2001research} report that non-native speakers tend to formulate in their native language and mentally translate to the target second language.
We assume that most crowdworkers have never written an academic paper, and that the target users of SentRev-based systems also include this type of inexperienced writers.

To control the quality of the drafts, we first create many candidate pairs of drafts and reference sentences ($X^{\text{cand}}$, $Y^{\text{cand}}$) and then filter them to create the quality-controlled set ($X$, $Y$). The following subsections detail this process.

\paragraph{Collecting final version sentences}
\label{subsec:collect_Y}
We collected sentences $Y^{\text{cand}}$ from the ACL Anthology Sentence Corpus (AASC).\footnote{\url{https://github.com/KMCS-NII/AASC}}
We extracted the sentences that satisfied the following conditions from the AASC as $Y^{\text{cand}}$:
\vspace{-0.2cm}
\begin{itemize}
 \setlength{\parskip}{0cm}
 \setlength{\itemsep}{0cm}
  \item accepted to ACL 2018,
  \item 70 to 120 characters long,
  \item does not include mathematical symbols, special tokens for citations, URLs, Greek letters, or other special symbols defined in AASC, and
  \item free of clear conversion mistakes when automatically extracted from PDFs.
\end{itemize}

\paragraph{Creating draft sentences}
\label{subsec:get_the_X_from_the_Y}

We used Japanese as $L'$.
First, we translated $Y^{\text{cand}}$ into Japanese using Google Translate.\footnote{\url{https://translate.google.com/}}
We denote the Japanese versions of $Y^{\text{cand}}$ by $Y^{\text{cand}}_{\text{ja}}$.
To guarantee the quality of $Y^{\text{cand}}_{\text{ja}}$, two authors of this paper, who were native speakers of Japanese, inspected all the sentences from $Y^{\text{cand}}_{\text{ja}}$ and removed those that at least one speaker judged to be incorrect translations.

Next, we asked each Japanese crowdworker to translate three sentences from $Y^{\text{cand}}_{\text{ja}}$ into English $Y^{\text{cand}}_{\text{ja}\to \text{en}}$ within 15 minutes.
The appropriate time limit and rules were determined based on several trial tasks.

The workers were allowed to insert the special symbol \texttt{<*>} in places where they could not think of a good expression for that position in their answer $Y^{\text{cand}}_{\text{ja}\to \text{en}}$.
This instruction revealed the information gaps that the authors of the drafts consciously left empty.
An author may also be unaware that a draft sentence is  missing sections.
306 workers participated in our crowdsourcing task.

\paragraph{Quality control}
\label{subsec:filtering_x_and_y}
We designed thorough filtering criteria and applied them to the workers because Yahoo! crowdsourcing, \footnote{\url{https://crowdsourcing.yahoo.co.jp/}} a Japanese crowdsourcing service, does not provide filtering based on the worker's writing skills or abilities.
We filtered workers depending on their writing activities.
We scored each worker using the three answers they produced by using the criteria detailed in Table~\ref{tbl:worker_quality}. We accepted work from workers with score 0 or higher as valid.
The hyperparameters were determined with trial experiments.
We used spaCy-CLD\footnote{\url{https://github.com/nickdavidhaynes/spacy-cld}} 
for language detection.

 \begin{table}[t]
     \centering
     {\footnotesize
     \begin{tabular}{p{50mm} l} \toprule
         Criteria  & Judgment \\ 
         \cmidrule(r){1-1} \cmidrule{2-2}
         Working time is too short ($<$ 2 minutes) & Reject \\
         All answers are too short ($<$ 4 words) & Reject \\ 
         No answer ends with ``.'' or ``?'' & Reject \\ 
         Contain identical answers & Reject \\ 
         Some answers have Japanese words & Reject \\
         No answer is recognized as English & Reject \\
         Some answers are too short ($<$ 4 words) & -2 points \\
         Some answers use fewer than 4 kinds of words & -2 points \\
         Too close to automatic translation (20 $<=$ L.D. $<=$ 30) & -0.5 points/ans \\
         Too close to automatic translation (10 $<=$ L.D. $<=$ 20) & -1.5 points/ans \\
         Too close to automatic translation (L.D. $<=$ 10) & Reject \\
         All answers end with ``.'' or ``?'' & +1 points\\
         Some answers have \textbf{\texttt{<*>}} & +1 points\\
         All answers are recognized as English & +1 points\\ \bottomrule
         \end{tabular}
     }
     \renewcommand{\arraystretch}{0.2}
     \caption{Criteria for evaluating workers. L.D denotes the Levenshtein distance.}
     \label{tbl:worker_quality}
 \end{table}

In addition, to remove instances with a too large gap, we automatically filtered out the obtained ($x^{\text{cand}}$, $y^{\text{cand}}$) $ \in (X^{\text{cand}}, Y^{\text{cand}})$ whose unigram overlap coefficient was considerably low:

\vspace{-0.2cm}
\begin{align}
\nonumber    \frac{|U(x^{\text{cand}}_{\text{checked}}) \cap U(y^{\text{cand}})|}{\text{min} \{|U(x^{\text{cand}}_{\text{checked}})|, |U(y^{\text{cand}})| \}} < \alpha \:\:,
\end{align}
where $U(\cdot)$ is the set of tokens excluding stop-words and special tokens (\texttt{<*>}). 
$x^{\text{cand}}_{\text{checked}}$ is the spell-checked version\footnote{We corrected spelling errors using \url{https://github.com/barrust/pyspellchecker}} of $x^{\text{cand}}$.
$\alpha$ is set to 0.4, which was determined in trial experiments.

We collected 10,804 pairs of draft and their final versions, which cost us approximately US\$4,200, including the trial rounds of crowdsourcing.

Unfortunately, works produced by unmotivated workers could have evaded the aforementioned filters and lowered the quality of our dataset. For example, workers could have bypassed the filter by simply repeating popular phrases in academic writing (``We apply we apply'').
To estimate the frequency of such examples, we sampled 100 $(x, y)$ pairs from $(X, Y)$ and asked an NLP researcher (not an author of this paper) fluent in Japanese and English to check for examples where $x$ was totally irrelevant to $x_{\text{ja}}$, which was shown to the crowdworkers when creating $x$.
The expert observed no completely inappropriate examples, but noted a small number of clearly subpar translations. 
Therefore, 95\% of sentence pairs were determined to be appropriate.
This result shows that, overall, our method was suitable to create the dataset and confirms the quality of \textsc{Smith}.

\subsection{Statistics}
Table~\ref{tab:comp_corpora} shows the statistics of our \textsc{Smith} dataset and a comparison with major datasets for building a writing assistance system \citep{napoles2017jfleg, mizumoto-etal-2011-mining, daudaravicius2015aesw}. 
The size of our dataset (10k sentence pairs) is six times greater than that of JFLEG, which contains both grammatical errors and nonfluent wording.
In addition, our dataset simulates significant editing---99\% of the pairs have some changes between the draft and its corresponding reference, and 33\% of the draft sentences contain gaps indicated by the special token \texttt{<*>}. 
We also measured the amount of change from the drafts $X$ to the references $Y$ by using the Levenshtein distance between them.
A higher Levenshtein distance between the $X$ and $Y$ sentences in our dataset indicated more significant differences between them compared with major GEC corpora.
This finding implies that our dataset emulates more drastic rephrasing.

\section{Analysis of the \textsc{Smith} dataset}
\label{sec:analysis_smis}
In this section, we run extensive analyses on the sentences written by non-native workers (\textit{draft} sentences $X$), and the original sentences extracted from the set of accepted papers (\textit{reference} sentences $Y$). 
We randomly selected a set of 500 pairs from \textsc{Smith} as the development set for analysis.

\begin{table}[t]
\centering
\renewcommand{\arraystretch}{1}
\small{
    \begin{tabular}{lcccc} \toprule
     Dataset & size & w/mask & w/change & L.D. \\
    \cmidrule(r){1-1} \cmidrule{2-2} \cmidrule{3-3} \cmidrule{4-4} \cmidrule{5-5}
    Lang-8 & 2.1M & - & 42\% & 3.5  \\
    AESW & 1.2M & - & 39\% & 4.8 \\
    JFLEG & 1.5k & - & 86\% & 12.4 \\
    \cmidrule(r){1-1} \cmidrule{2-2} \cmidrule{3-3} \cmidrule{4-4} \cmidrule{5-5}
    \textsc{Smith} & 10k & 33\% & 99\% & 47.0 \\ \bottomrule
    \end{tabular}
    \caption{Comparison with existing datasets. w/mask and w/change denote the percentage of source sentences with mask tokens and the percentage where the source and target sentences differ, respectively. L.D. indicates the averaged character-level Levenshtein distance between the pairs of sentences.}
    \label{tab:comp_corpora}
}
\end{table}

\begin{figure}[t]
    \centering
      \includegraphics[width=7.5cm]{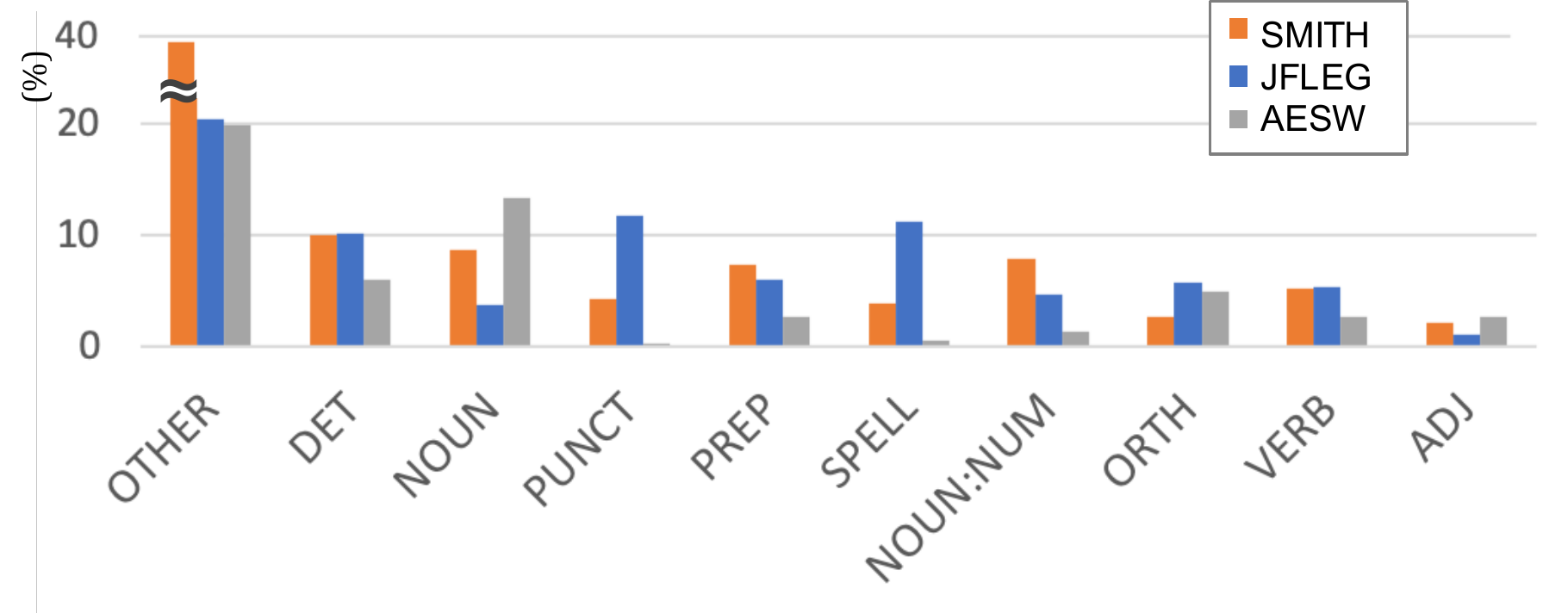}
      \vspace{-0.1cm}
      \caption{Comparison of the top 10 frequent errors observed in the 3 datasets.}
      \label{fig:anno_errortype}
\end{figure}

\begin{figure}[t]
    \centering
      \includegraphics[width=7.5cm]{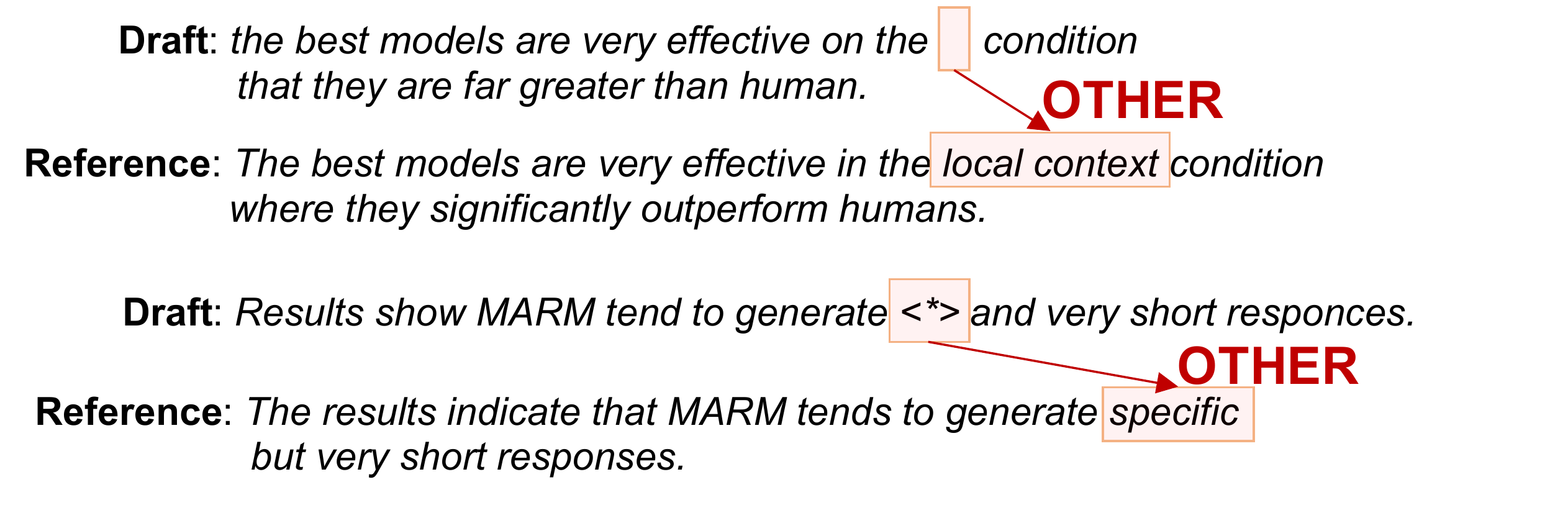}
      \vspace{-0.1cm}
      \caption{Examples of ``OTHER'' operations predicted by the ERRANT toolkit.}
      \label{fig:other_examples}
\end{figure}

\subsection{Error type comparison}
To obtain the approximate distributions of error types between the source and target sentences, we used ERRANT~\citep{bryant2017automatic, felice-etal-2016-automatic}.
Next, we compared them with three datasets: \textsc{Smith}, AESW (the same domain as \textsc{Smith}), and JFLEG (has a relatively close Levenshtein distance to \textsc{Smith}).
To calculate the error type distributions on AESW and JFLEG, we randomly sampled 500 pairs of source and target sentences from each corpus.
Figure~\ref{fig:anno_errortype} shows the results of the comparison.
Although all datasets contained a mix of error types and operations, the \textsc{Smith} dataset included more ``OTHER'' operations than the other two datasets.
Manual inspection of some samples of ``OTHER'' operations revealed that they tend to inject information missing in the draft sentence (Figure~\ref{fig:other_examples}). This finding confirms that our dataset emphasizes a new, challenging ``completion-type'' task setting for writing assistance.

\subsection{Human error type analysis}
\label{subsec:error_type_analysis}

\begin{figure}[t]
    \centering
      \includegraphics[width=7.5cm]{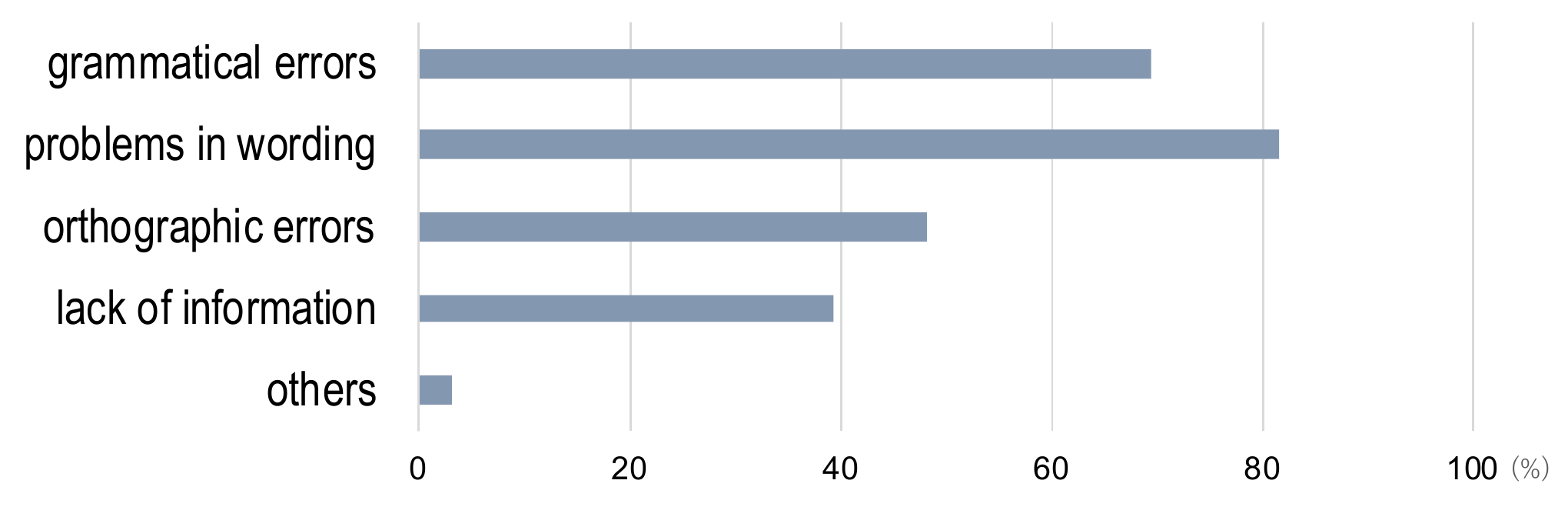}
      \vspace{-0.1cm}
      \caption{Result of the English experts' analyses of error types in draft sentences on our \textsc{Smith} dataset. The scores show the ratio of sentences where the targeted type of errors occurred.}
      \label{fig:anno_errortype_expert}
\end{figure}

To understand the characteristics of our dataset in detail, an annotator proficient in English (not an author of this paper) analyzed the types of errors in the draft sentences (Figure~\ref{fig:anno_errortype_expert}). 
The most frequent errors were \textit{fluency problems} (e.g.,  ``In these \emph{ways}'' instead of ``In these \emph{methods},'')---characterized by errors in academic style and wording, which are out of the scope of traditional GEC. Another notable type of frequent error was \textit{lack of information}, which further distinguishes this dataset from other datasets. 

\subsection{Human fluency analysis}
\label{subsec:fluency-analysis}

We outsourced the scoring of the fluency of the given draft and reference sentence pairs to three annotators proficient in English. Nearly every draft $x$ (94.8\%) was marked as being less fluent than its corresponding reference $y$, 
confirming that obtaining high performance with our dataset requires the ability to transform rough input sentences into more fluent sentences.

\subsection{Sentence-level linguistic characteristics}
\label{subsub:linguistic_char}

\begin{table}[t]
     \centering
     \renewcommand{\arraystretch}{1.0}
     \footnotesize{
     \begin{tabular}{lcp{13mm}p{14mm}c} \toprule
      Data  & FRE &  passive voice (\%) & word repetition (\%) & PPL \\ 
      \cmidrule(lr){1-1}  \cmidrule(lr){2-4}  \cmidrule(r){5-5}
      Draft $X$ & 45.5 & \multicolumn{1}{c}{34.0} & \multicolumn{1}{c}{33.0}& 1373 \\
      Reference $Y$ & 40.0 & \multicolumn{1}{c}{29.6} & \multicolumn{1}{c}{28.6} & 147 \\\bottomrule
     \end{tabular}
     }
         \caption{Comparison of the draft and reference sentences in \textsc{Smith}. FRE and PPL scores were calculated once in each sentence and then averaged over all the sentences in the development set of \textsc{Smith}.}
         \label{tab:sent_analysis}
\end{table}

We computed some sentence-level linguistic measures over the dataset sentences: Flesch Reading Ease (FRE)~\citep{flesch1948reading_ease}, passive voice\footnote{\url{https://github.com/armsp/active_or_passive}}, word repetition, and perplexity (PPL) (Table~\ref{tab:sent_analysis}).

FRE measures the \emph{readability} of a text, namely, how easy it is to understand (higher is easier). 
The draft sentences consistently demonstrated higher FRE scores than their reference counterparts, which may be attributed to the latter containing more sophisticated language and technical terms. 

In addition, workers tended to use the passive voice and to repeat words within a narrow span, and  both those phenomenon must be avoided in academic writing. 
We conducted further analyses on lexical tendencies between the drafts and references (Appendix~\ref{app:lexical_tendencies}).

Finally, we analyzed the draft and the reference sentences using PPL calculated by a 5-gram language model trained on ACL Anthology papers.\footnote{PPL is calculated with the implementation available in the KenLM (\url{https://github.com/kpu/kenlm}), tuned on AASC (excluding the texts used for building the \textsc{Smith}).}
The higher PPL scores in the draft sentences (Table~\ref{tab:sent_analysis}) suggest that they have properties unsuitable for academic writing (e.g., less fluent wording).

\section{Experiments}
\label{sec:exp}

 \begin{table*}[t]
    \centering
    \label{tab:ex_inc}
    \footnotesize{
    \begin{tabularx}{16cm}{lXX} 
    \toprule
      method & original & generated\\ 
      \midrule
      Heuristic &
        Besides , the recognizer successfully rejected only 15 out of 42 negative sentences .& 
        recognizer Besides successfully , the informativeness rejected of out \texttt{<*>} \\ 
        
      \midrule 
      Grammatical error generation & 
        We plan to \textbf{analyze} these direct communications \textbf{and} interaction of sentiments \textbf{expressed} in these sequences of posts . & 
        We plan to \textbf{analysis} the direct communication interaction of sentiments \textbf{express} in these sequence of posts .\\ 
      \midrule
      Style removal &
        This experiment \textbf{suggested} that there were ambiguities in these pointing gestures and \textbf{led to a redesign} of the system . & 
        This experiment \textbf{indicated} the ambiguity found in the pointing gestures and \textbf{caused a renewal} of the system .\\
      \midrule
      Entailed sentence generation & 
        Figure 2 \textbf{illustrates the effectiveness} of different features class. & 
        There is different feature in figure 2 .\\
      \bottomrule
    \end{tabularx}
    }
    \caption{Examples of generated training dataset.}
    \label{tbl:generated_train_data}
\end{table*}

\subsection{Baseline models}
\label{subsec:exp}
We evaluated three baseline models on the SentRev task.

\subsubsection{Heuristic noising and denoising model}
\label{subsubsec:heuristic}
We can access a great deal of final version academic papers.
Noising and denoising approaches have gained attention in the GEC and machine translation fields~\citep{edunov2018understanding, xie-etal-2018-noising, lichtarge-etal-2019-corpora}.
We combined these two factors to train baseline models on noised final version sentences.

First, we collected 4,898,146 sentences $Y^{\text{aasc}}$ from the AASC that satisfied the following conditions: (i) not included in the \textsc{Smith} dataset, (ii) not too long or too short (between 5 and 35 tokens), (iii) over 50\% of the characters were alphabetic.
Next, we created a training dataset $(X^{\text{aasc}}_{\text{hrst}}, Y^{\text{aasc}})$ by adding noise to $Y^{\text{aasc}}$.

As the simplest approach for noising, we used a set of heuristic rules by randomly deleting, replacing, and swapping words in the reference sentences. 
Specifically, these rules included deleting words with a probability of 0.1, replacing words with a token that appeared over 10,000 times in $Y_{\text{aasc}}$ with a probability of  0.1, and randomly shuffling the sentence while maintaining the originally adjacent words within three words apart. 
Next, we randomly replaced up to 50$\%$ of the words with a \texttt{<*>} token (see Appendix~\ref{app:masking_algo} for a more detailed algorithm).
This method generated 4.8M heuristically noised sentences.
 
Subsequently, we trained a denoising model (a mapping function from $X^{\text{aasc}}_{\text{hrst}}$ to $Y^{\text{aasc}}$) by using  Transformer~\citep{vaswani2017transformer} implemented in fairseq~\citep{ott2019fairseq}.
We used an Adam optimizer~\citep{Kingma2015adam} with $\alpha = 0.0005$, $\beta_{1} = 0.9$, $\beta_{2} = 0.98$, and $\epsilon = 10e^{-8}$.
We limited the maximum tokens per each minibatch to 3000, limited the maximum number of updates to 500,000, and used a dropout rate of 0.3.
The input and output texts were tokenized and then segmented into character bigrams. We used a beam width of 5 in the decoding.
This model is our first baseline model for the SentRev task (henceforth, H-ND).

\subsubsection{Enc-Dec noising and denoising model}

\label{subsubsec:enc-dec}
As an extension of the heuristic noising and denoising model, we changed the noising methods to better simulate the characteristics of $X$ in \textsc{Smith} than the heuristic rules in Section~\ref{subsubsec:heuristic}.
As described in Section~\ref{sec:analysis_smis}, the drafts tended to (i) contain grammatical errors, (ii) use stylistically improper wording, and (iii) lack certain words. 
We used the following three neural Encoder-Decoder (Enc-Dec) models to generate the synthetic draft sentences.

\begin{table*}[t]
    \centering
    \renewcommand{\arraystretch}{1.0}
    \small{
    \begin{tabular}{lcccccccccc} \toprule
      Model & BLEU & ROUGE-L &  BERT-P & BERT-R & BERT-F &  P & R & F$_{0.5}$ & Gramm. & PPL  \\ 
      \cmidrule(lr){1-1}  \cmidrule(lr){2-2}  \cmidrule(lr){3-3} \cmidrule(lr){4-6} \cmidrule(lr){7-9} \cmidrule(lr){10-10} \cmidrule(lr){11-11}
      Draft $X$ & 9.8 & 46.8 & 75.9 & 78.2 & 77.0 & - & - & - & 92.9 & 1454\\
      \cmidrule(lr){1-1}  \cmidrule(lr){2-2}  \cmidrule(lr){3-3} \cmidrule(lr){4-6} \cmidrule(lr){7-9} \cmidrule(lr){10-10} \cmidrule(lr){11-11}
      H-ND & 8.2 & 45.0 & 77.0 & 76.1 & 76.5 & 5.4 & 2.9 & 4.6 & 94.1 & 406 \\
      ED-ND & {\bf 15.4} & {\bf 51.1} & {\bf 80.9} & {\bf 80.0} & {\bf 80.4} & 21.8 & {\bf 12.8} & {\bf 19.2} & 96.3 & {\bf 236}\\
      GEC & 11.9 & 49.0 & 80.8 & 79.1 & 79.9 & {\bf 22.2} & 6.2 & 14.6 & {\bf 96.7} & 414 \\
      \cmidrule(lr){1-1}  \cmidrule(lr){2-2}  \cmidrule(lr){3-3} \cmidrule(lr){4-6} \cmidrule(lr){7-9} \cmidrule(lr){10-10} \cmidrule(lr){11-11}
      Reference $Y$ & - & - & - & - & - & - & - & - & 96.5 & 147 \\ \bottomrule
    \end{tabular}
    }
    \caption{Results of quantitative evaluation. Gramm. denotes the grammaticality score.}
    \label{tab:result}
\end{table*}

\begin{table*}[t]
    \centering
    \renewcommand{\arraystretch}{0.5}
    \small{
        \begin{tabularx}{16cm}{lX}
        \toprule
        Draft & The global modeling using the reinforcement learning in all documents is our work in the future .  \\ 
        \cmidrule(lr){1-1}  \cmidrule(lr){2-2}
        H-ND & The global modeling \textbf{of} the reinforcement learning \textbf{using} all documents \textbf{in} our work \textbf{is} the future . \\
        \cmidrule(lr){1-1}  \cmidrule(lr){2-2}
        ED-ND & \textbf{In our future work , we plan to explore the use of} global modeling \textbf{for} reinforcement learning in all documents .    \\ 
        \cmidrule(lr){1-1}  \cmidrule(lr){2-2}
        GEC & Global modelling using reinforcement learning in all documents is our work in the future .\\ 
        \cmidrule(lr){1-1}  \cmidrule(lr){2-2}
        Reference & The global modeling using reinforcement learning for a whole document is our future work .\\
        \midrule
        Draft & Also , the above \textbf{\texttt{<*>}} efficiently calculated by dynamic programming . \\
        \cmidrule(lr){1-1}  \cmidrule(lr){2-2}
        H-ND & Also , the above \textbf{results are calculated} efficiently by dynamic programming . \\
        \cmidrule(lr){1-1}  \cmidrule(lr){2-2}
        ED-ND & Also , the above \textbf{probabilities are calculated} efficiently by dynamic programming . \\
        \cmidrule(lr){1-1}  \cmidrule(lr){2-2}
        GEC & Also , the above {\bf is} efficiently calculated by dynamic programming .\\ 
        \cmidrule(lr){1-1}  \cmidrule(lr){2-2}
        Reference & Again , the above equation can be efficiently computed by dynamic programming . \\
        \midrule
        Draft & Chart4 : relation model and gold \% between KL and piason . \\
        \cmidrule(lr){1-1}  \cmidrule(lr){2-2}
        H-ND & \textbf{Table 1 : Charx-} relation between \textbf{gold and piason and KL} .\\
        \cmidrule(lr){1-1}  \cmidrule(lr){2-2}
        ED-ND & \textbf{Figure 2 : CharxDiff} relation between \textbf{model and gold standard and piason} . \\
        \cmidrule(lr){1-1}  \cmidrule(lr){2-2}
        GEC & Chart4 : relation model and gold \% between KL and person . \\ 
        \cmidrule(lr){1-1}  \cmidrule(lr){2-2}
        Reference & Table 4 : KL and Pearson correlation between model and gold probability . \\
        \bottomrule
        \end{tabularx}
    }
\caption{Examples of the output from the baseline models. Bold text indicates tokens introduced by the model.}
\label{tbl:generation}
\end{table*}

\paragraph{Grammatical error generation}
Here, we trained a model that introduces synthetic grammatical errors to ``clean'' sentences by using a ``flipped'' dataset from GEC (clean $\to$ erroneous). We used nonidentical (source, target) sentence pairs from the Lang-8, AESW, and JFLEG datasets.

\paragraph{Style removal} 
To generate stylistically unnatural sentences in the academic domain, we used paraphrasing, which preserves a sentence's content while disregarding its style.
We used the ParaNMT-50M dataset~\citep{wieting-gimpel-2018-paranmt}, a paraphrase dataset automatically created using Enc-Dec translation.
We extracted parallel sentences with annotated paraphrase scores between 0.7 and 0.95 from the ParaNMT-50M dataset and used swapped pairs of source and target sentences in the dataset.

\paragraph{Entailed sentence generation}
To simulate the missing words in the draft sentences, we trained a model that generated a sentence entailed with the given text.
We extracted entailed sentence pairs from the SNLI~\citep{bowman2015snli} and the MultiNLI~\citep{williams2018multinli} datasets.

\paragraph{Random noising beam search}
As \citet{xie-etal-2018-noising} pointed out, a standard beam search often yields hypotheses that are too conservative. 
This tendency leads the noising models to generate synthetic draft sentences similar to their references.
To address this problem, we applied the random noising beam search \cite{xie-etal-2018-noising} on all three noising models.
Specifically, during the beam search, we added $r\beta$ to the scores of the hypotheses, where $r$ is a value sampled from a uniform distribution over the interval $[0,1]$, and $\beta$ is a penalty hyperparameter set to 5.

\paragraph{}
We obtained 14.6M sentence pairs of ($X^{\text{aasc}}_{\text{encdec}}$, $Y^{\text{aasc}}$) by applying these Enc-Dec noising models to $Y^{\text{aasc}}$.
To train the denoising model, we used both data ($X^{\text{aasc}}_{\text{hrst}}$, $Y^{\text{aasc}}$) and ($X^{\text{aasc}}_{\text{encdec}}$, $Y^{\text{aasc}}$).
The model architecture was the same as the heuristic model.
This denoising model is our second baseline model (ED-ND).
To facilitate research in the SentRev task, we released all the 19.6M synthetic data.\footnote{\url{https://github.com/taku-ito/INLG2019_SentRev}}

\begin{figure}[t]
    \centering
      \includegraphics[width=7.5cm]{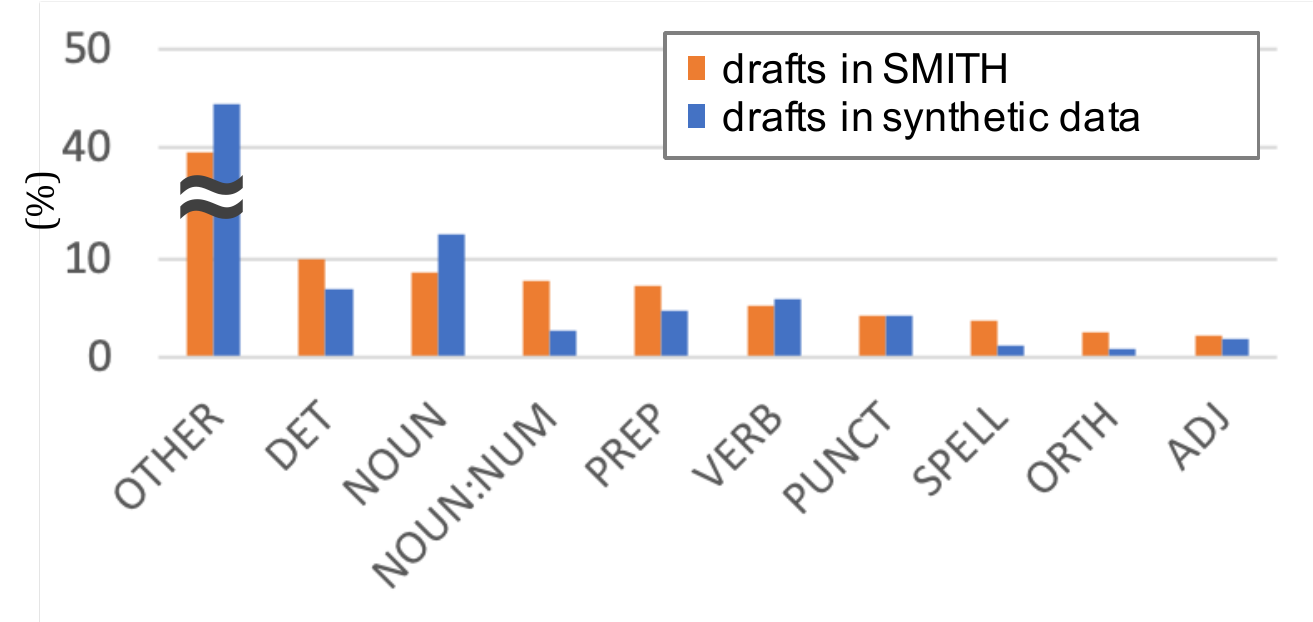}
      \vspace{-0.1cm}
      \caption{Comparison of the 10 most frequent error types in \textsc{Smith} and synthetic drafts created by the Enc-Dec noising methods.}
      \label{fig:error_types_synthetic_dataset}
\end{figure}

\paragraph{Analysis of the synthetic drafts}
Finally, we analyzed the error type distribution of the synthetic data used for training Enc-Dec noising and denoising model with ERRANT  (Figure~\ref{fig:error_types_synthetic_dataset}).
The error type distribution from the synthetic dataset had similar tendencies to the one from the development set in \textsc{Smith} (real-draft).
Kullback–Leibler divergence between these error type distributions was 0.139.
This result supports the validity of our assumption that the SentRev task is a combination of GEC, style transfer, and a completion-type task.

\begin{figure}[t]
    \centering
      \includegraphics[width=7.5cm]{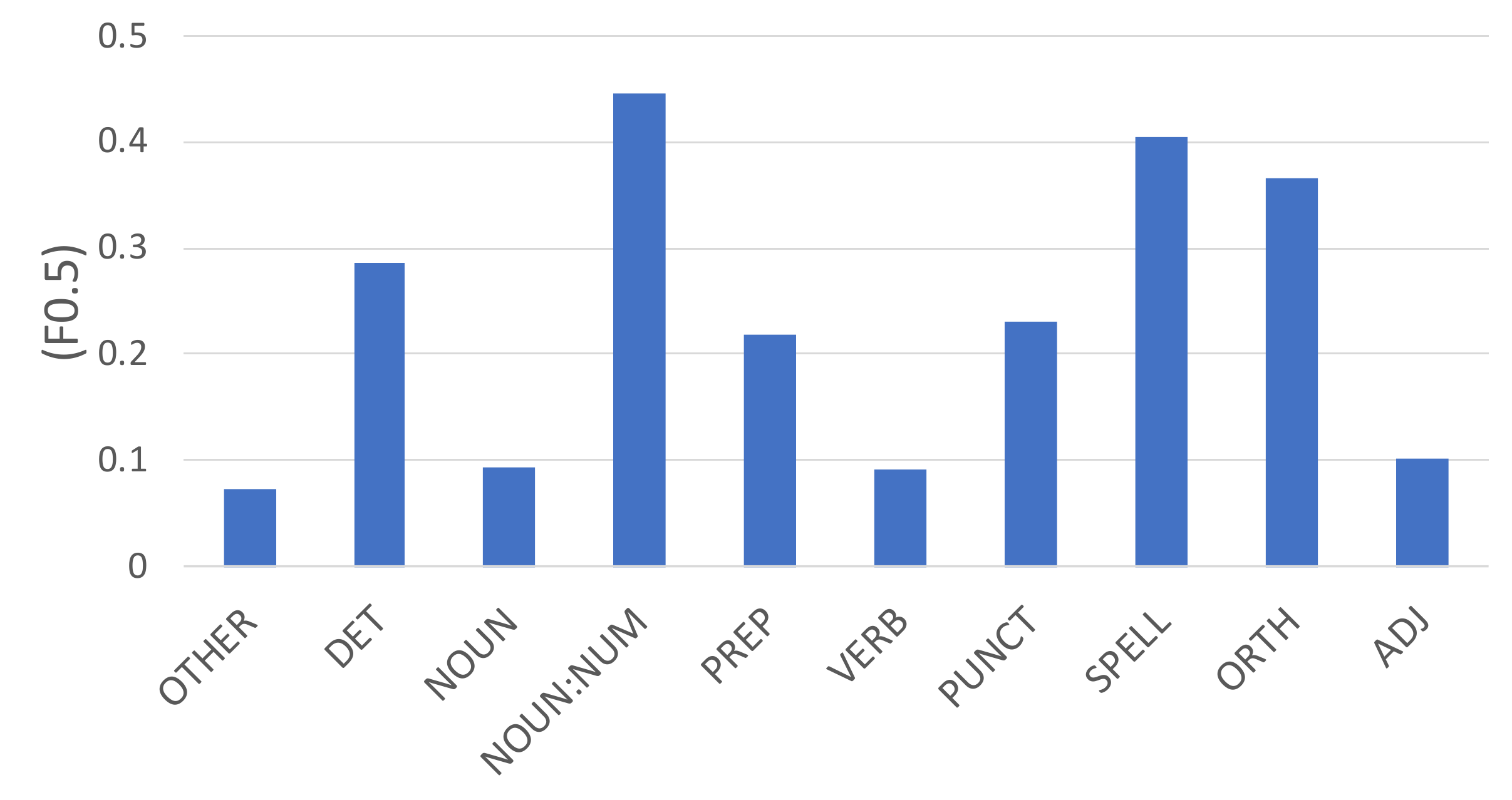}
      \vspace{-0.1cm}
      \caption{Performance of the ED-ND baseline model on top 10 most error types in SMITH.}
      \label{fig:performance_error_types}
\end{figure}

Table~\ref{tbl:generated_train_data} shows examples of the training data generated by the noising models described in Section~\ref{sec:exp}.
Heuristic noising, the rule-based noising method, created ungrammatical sentences.
The grammatical error generation model added grammatical errors (e.g., {\it plan to analyze} $\to$ {\it plan to analysis}).
The style removal model generated stylistically unnatural sentences for the academic domain (e.g., {\it redesign} $\to$ {\it renewal}). 
The entailed sentence generation model caused a lack of information.

\subsubsection{GEC model} 
\label{subsubsec:gec}
The GEC task is closely related to SentRev.
We examined the performance of the current state-of-the-art GEC model~\citep{zhao-etal-2019-improving} in our task.
We applied spelling correction before evaluation following \citet{zhao-etal-2019-improving}.

\subsection{Evaluation metrics}
The SentRev task has a very diverse space of valid revisions to a given context, which is challenging to evaluate.
As one solution, we evaluated the performance from multiple aspects by using various reference and reference-less evaluation metrics.
We used BLEU, ROUGE-L, and F$_{0.5}$ score, which are widely used metrics in related tasks (machine translation, style-transfer, GEC).
We used nlg-eval~\citep{sharma2017nlgeval} to compute the BLEU and ROUGE-L scores and calculated F$_{0.5}$ scores with ERRANT.
In addition, to handle the lexical and compositional diversity of valid revisions, we used BERT-score~\citep{bert-score}, a contextualized embedding-based evaluation metric.
Furthermore, we used two reference-less evaluation metrics: grammaticality score~\citep{napoles2016referenceless} and PPL.
Grammaticality was scored as $1-({N_{\text{errors}\:\text{in}\:\text{sentence}}}/{N_{\text{tokens}\:\text{in}\:\text{sentence}}})$, where the number of grammatical errors in a sentence is obtained using LanguageTools.\footnote{\url{https://github.com/languagetool-org/languagetool/releases/tag/v3.2}}
By using a language model tuned to the academic domain, we expect PPL to evaluate the stylistic validity and fluency of a complemented sentence.
We favored n-gram language models over neural language models for reproducibility and calculated the score in the same manner as described in Section~\ref{subsec:fluency-analysis}.

\section{Results}
Table~\ref{tab:result} shows the performance of the baseline models.
We observed that the ED-ND model outperforms the other models in nearly all evaluation metrics.
This finding suggests that the Enc-Dec noising methods induced noise closer to real-world drafts compared with the heuristic methods.

The current state-of-the-art GEC model showed higher precision but low recall scores in F$_{0.5}$.
This suggests that the SentRev task requires the model to make a more drastic change in the drafts than in the GEC task.
Furthermore, the GEC model, trained in the general domain, showed the worst performance in PPL.
This indicates that the general GEC model did not reflect academic writing style upon revision and that SentRev requires academic domain-aware rewriting.

Table~\ref{tbl:generation} shows examples of the models' output.
In the first example, the ED-ND model made a drastic change to the draft. 
The middle example demonstrates that our models replaced the \texttt{<*>} token with plausible words.
The last example is the case where our model underperformed by making erroneous edits such as changing {\it ``Chart4''} to {\it ``Figure2''}, and suggesting odd content ({\it ``relation between model and gold standard and piason''}). 
This may be due to  having inadvertently introduced noise while generating the training datasets.
Appendix~\ref{app:example_gen_smith} shows more examples of generated sentences.
Using ERRANT, we analyzed the performance of the ED-ND baseline model by error types.
The results are shown in Figure~\ref{fig:performance_error_types}.
Overall, typical grammatical errors such as noun number errors or orthographic errors are well corrected, but the model struggles with drastic revisions (``OTHER" type errors).

\section{Related work}
\label{sec:related_work}

\subsection{Writing assistance in the academic domain}
Several shared tasks for assisting academic writing have been organized.
The Helping Our Own (HOO) 2011 Pilot Shared Task~\citep{dale2011helping} aimed to promote the development of tools and techniques to assist authors in writing, with a specific focus on writing within the NLP community. 
The Automated Evaluation of Scientific Writing (AESW) Shared Task~\citep{daudaravicius2015aesw} was organized to promote tools to help write scientific papers.
The HOO dataset was created by finding errors in published papers and editing the errors, and the AESW dataset contains a collection of text extracts from published journal articles before and after proofreading. Rather than adding finishing touches to almost completed sentences, our task is to convert unfinished, rough drafts into complete sentences.
In addition, these studies tackled the task of the \textit{identification} of errors while SentRev goes further by \textit{rewriting} the drafts.

Other corpora for revisions are available in the academic domain \citep{lee-webster-2012-corpus, tan-lee-2014-corpus, zhang-etal-2017-corpus}. 
Thus, we provide a notable contribution by exploring the methods to create a dataset of revisions with a scalable crowdsourcing approach. 
By contrast, \citet{zhang-etal-2017-corpus} recruited 60 students over 2 weeks and \citet{lee-webster-2012-corpus} collected data from a language learning project where over 300 tutors reviewed academic essays written by 4500 students.

\subsection{Grammatical error correction}
GEC is the task of correcting errors in text such as spelling, punctuation, grammar, and word choice~\citep{ng2014, yuan2016}. 
GEC falls within the \textit{editing} and \textit{proofreading} phases of the writing process, while SentRev subsumes GEC and a broader range of text generation (e.g., increasing the fluency of the sentence and complementing missing information).
\citet{napoles2017jfleg} and \citet{sakaguchi2016reassessing} explored fluency edits to correct grammatical errors and to make a text more ``native sounding.''
Although this direction is similar to SentRev, our task used sentences that required many more corrections.

\subsection{Style transfer}
Style transfer is the task of rephrasing the text to conform to specific stylistic properties while preserving the text's original semantic content~\citep{logeswaran2018content_preserving_generation, prabhumoye2018style_transfer}. 
From the perspective of automatic academic writing assistance, the assistance systems are required to convert nonacademic-style drafts into academic-style drafts.
This type of transfer is regarded as a subproblem in the \textit{revising} stage of the writing process.

\subsection{Text completion}
The drafts in the \textit{revising} stage may contain gaps denoted with \texttt{<*>}. 
This setting is similar to \textit{text infilling}~\cite{wanrong2019textinfilling}, masking-based language modeling~\cite{fedus2018maskgan, devlin:2018:bert}, or the \textit{sentence completion task}~\cite{Zweig2012}, where the models are required to replace mask tokens with plausible words. 
Notably, SentRev differs from such tasks because systems for these tasks are expected to keep all the original tokens unchanged and only fill the \texttt{<*>} token, with one or more other tokens.

\section{Conclusion and future work}
We proposed the SentRev task, where an incomplete, rough draft sentence is transformed into a more fluent, complete sentence in the academic writing domain. 
We created the \textsc{Smith} dataset with crowdsourcing for development and evaluation of this task and established baseline performance with a synthetic training dataset. 
We believe that this task can increase the effectiveness of  the process of academic writing.
In future work, we plan to improve the information gap-filling aspect of revision by considering the surrounding context of target sentences.
In addition, to develop a more holistic writing assistance tool, we plan to extend our system to be able to suggest diverse correction candidates, provide interactive assistance, and integrate translation systems. 

\section{Acknowledgements}
We thank the Tohoku NLP laboratory members who provided us with their valuable advice. 
We are grateful to Benjamin Heinzerling and Marie-Jos\'{e}e Brassard for their feedback.
We are also grateful to Masato Mita for advice on the experiments.
The work of J. Suzuki was partly supported by JSPS KAKENHI Grant Number 19H04162.

\bibliographystyle{acl_natbib}
\bibliography{acl2019}

\clearpage

\appendix

\begin{figure*}[t]
    \centering
      \includegraphics[width=12cm]{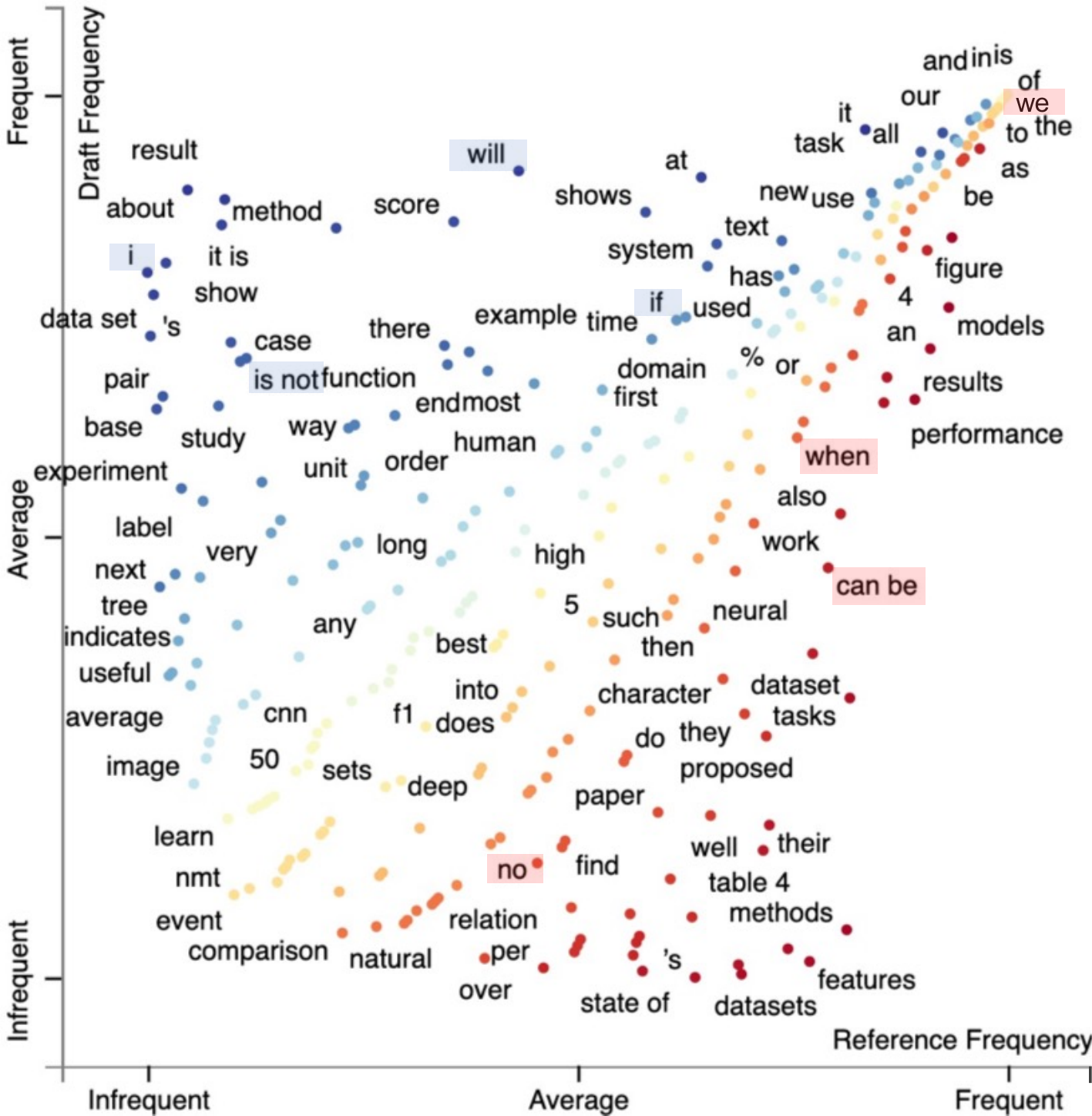}
      \vspace{-0.1cm}
      \caption{Characteristic words and phrases in draft sentences and reference sentences in the development set of \textsc{Smith}.}
      \label{fig:word_ana}
\end{figure*}

\section{Lexical tendencies}
\label{app:lexical_tendencies}

Certain words and phrases were more frequently observed in the reference sentences than in the draft sentences, and vice-versa.
Figure~\ref{fig:word_ana} visualizes these biases, where 
words more often observed in the draft sentences are plotted in the upper-left corner, and words more often observed in the references are plotted in the lower-right corner. Words observed more commonly in the drafts were: \textit{will}, \textit{is not}, \textit{if}, and \textit{I}, versus \textit{can be}, \textit{no}, \textit{when}, and \textit{they}. The contrast also includes a widely-used spelling (\textit{data set} vs \textit{dataset}) and common plurality (\textit{method} vs \textit{methods}).
The plot was generated using the scattertext toolkit~\citep{kessler2017scattertext}.

\section{Heuristic noising algorithm}
\label{app:masking_algo}
Algorithm~\ref{alg:alg_hueristic} shows the noising algorithm in the heuristic noising method.

\begin{algorithm*}[t]
\caption{Heuristic noising}
\label{alg:alg_hueristic}
\begin{flushleft}
\textbf{INPUT:} $x= \{w_0, w_1, \cdots, w_n\}$
\end{flushleft}
\begin{algorithmic}[1]
\STATE $x = \text{delete}(x, 0.1)$ \\
\# 10\% of the tokens in x are deleted. \\

\STATE $x = \text{replace}(x, 0.1)$ \\
\# 10\% of the tokens in x are replaced with common terms in ACL. \\

\STATE $x = \text{permutate}(x)$ \\
\# permutate the tokens in x. \\

\STATE $r \leftarrow \text{Uniform}(0, 0.5)$
\STATE $m = \text{int}(x\text{.length} * r)$
\STATE $c = 0$
\WHILE{$c<m$}
\STATE $n \leftarrow \text{sample}(\{j \in \mathcal{N}\;|\; 1 \leq j \leq m-c\})$
\STATE $(s, e) \leftarrow \text{sample}(\{n\text{-grams of } x\})$
\STATE $x = ``x_{:s-1} + \textbf{\texttt{<*>}} + x_{e+1:} \;''$
\STATE $c = c + n$
\ENDWHILE \\
\# $r\times100$\% of the tokens in x are masked. \\
\end{algorithmic}
\end{algorithm*}

\section{Examples from the \textsc{Smith} dataset and generated sentences by Baseline models}
\label{app:example_gen_smith}
Table~\ref{tbl:generation_2} shows examples from the \textsc{Smith} dataset and the output of the baseline models.
``Reference'' is a sentence extracted from papers, ``Draft'' is written by a crowdworker and is the input for the baseline models.

\begin{table*}[t]
    \centering
    \renewcommand{\arraystretch}{0.9}
    \small{
        \begin{tabularx}{16cm}{lX}
        \toprule
        Draft &
            By this setting , the persona is acquired from a test set popl about both turker anad model .\\ 
        \cmidrule(lr){1-1}  \cmidrule(lr){2-2}
        H-ND &
            By this setting , the persona is acquired from a test set both about popl anad anad model . \\ 
        \cmidrule(lr){1-1}  \cmidrule(lr){2-2}
        ED-ND &
            In this setting , persona is obtained from the test set popl about both Turker and model . \\ 
        \cmidrule(lr){1-1}  \cmidrule(lr){2-2}
        GEC &
            By this setting , the persona is acquired from a test set pool about both turkey and models . \\
        \cmidrule(lr){1-1}  \cmidrule(lr){2-2}
        Reference & 
            In this setting , for both the Turker and the model , the personas come from the test set pool .\\ \midrule
        Draft & 
            In addition to results of study until now , we add two baseline to vindicate effectiveness on our flame work . \\
        \cmidrule(lr){1-1}  \cmidrule(lr){2-2}
        H-ND &
            In addition to the results of this study , we now add two baseline methods to vindicate effectiveness on our work .\\
        \cmidrule(lr){1-1}  \cmidrule(lr){2-2}
        ED-ND &
            In addition to the results of the study until now , we add two baselines to visualize the effectiveness of our framework . \\
        \cmidrule(lr){1-1}  \cmidrule(lr){2-2}
        GEC &
            In addition to the results of study until now , we added two baseline to vindicate effectiveness on our flame work . \\
        \cmidrule(lr){1-1}  \cmidrule(lr){2-2}
        Reference &  
            In addition to results of previous work , we add two baselines to demonstrate the effectiveness of our framework . \\ \midrule
        Draft & 
            Yhe input and output \texttt{<*>} are one - hot encoding of the center word and the context word , \texttt{<*>} .	\\
        \cmidrule(lr){1-1}  \cmidrule(lr){2-2}
        H-ND & 
            The input and output are one - hot encoding of the center word and the context word , respectively . \\
        \cmidrule(lr){1-1}  \cmidrule(lr){2-2}
        ED-ND &
            The input and output layers are one - hot encoding of the center word and the context word , respectively . \\
        \cmidrule(lr){1-1}  \cmidrule(lr){2-2}
        GEC &
            Yhe input and output are one - hot encoding of the center word and the context word , . \\
        \cmidrule(lr){1-1}  \cmidrule(lr){2-2}
        Reference &  
            The input and output layers are centre word and context word one - hot encodings , respectively . \\ \midrule
        Draft & 
            I registered the vocabulary sizes of encorder and decorder as 150 K and 50 K each other . \\
        \cmidrule(lr){1-1}  \cmidrule(lr){2-2}
        H-ND & 
            I registered the vocabulary sizes of decorder and encorder as 150 K and each other . \\
        \cmidrule(lr){1-1}  \cmidrule(lr){2-2}
        ED-ND &
            We registered the vocabulary sizes of the encoder and decoder as 150 K and 50 K respectively . \\
        \cmidrule(lr){1-1}  \cmidrule(lr){2-2}
        GEC &
            I registered the vocabulary sizes of encoder and recorder as 150 K and 50 K for each other . \\
        \cmidrule(lr){1-1}  \cmidrule(lr){2-2}
        Reference &  
            In this experiment , we set the vocabulary size on the encoder and decoder sides to 150 K and 50 K , respectively . \\ \midrule
        Draft & 
            They add the new class image generated by generator and classfy them . \\
        \cmidrule(lr){1-1}  \cmidrule(lr){2-2}
        H-ND & 
            They add the new image class generated by the generator and classfy them .\\
        \cmidrule(lr){1-1}  \cmidrule(lr){2-2}
        ED-ND &
            They add a new class of images generated by the generator and classify them . \\
        \cmidrule(lr){1-1}  \cmidrule(lr){2-2}
        GEC &
            They add a new class image generated by generator and classify them . \\
        \cmidrule(lr){1-1}  \cmidrule(lr){2-2}
        Reference &  
            They add a new class of images that are generated by the generator and classify them . \\ \midrule
        Draft & 
            The chart 3 shows performance of multi input correction against sub groups with different number of witnesses .	 \\
        \cmidrule(lr){1-1}  \cmidrule(lr){2-2}
        H-ND & 
            Table 3 shows the performance of multi - chart correction against different input groups with different number of witnesses .	\\
        \cmidrule(lr){1-1}  \cmidrule(lr){2-2}
        ED-ND &
            Figure 3 shows the performance of multiple input correction against subgraphs with different number of witnesses . \\
        \cmidrule(lr){1-1}  \cmidrule(lr){2-2}
        GEC &
            chart 3 shows performance of multi input correction against sub groups with different number of witnesses . \\
        \cmidrule(lr){1-1}  \cmidrule(lr){2-2}
        Reference &  
            Figure 3 presents the performance of multi - input correction on subgroups with different number of witnesses . \\ \midrule
        Draft & 
            It is vindicated that InferSent accomplishes the most \texttt{<*>} result regarding SentEval task .  \\
        \cmidrule(lr){1-1}  \cmidrule(lr){2-2}
        H-ND & 
            It is vindicated that InferSent accomplishes the most relevant result regarding the SentEval task . \\
        \cmidrule(lr){1-1}  \cmidrule(lr){2-2}
        ED-ND &
            It is vindicated that InferSent accomplishes the most important result regarding the SentEval task . \\
        \cmidrule(lr){1-1}  \cmidrule(lr){2-2}
        GEC &
            It is vindicated that InferSent accomplishes the most results regarding SentEval task . \\
        \cmidrule(lr){1-1}  \cmidrule(lr){2-2}
        Reference & 
            InferSent has been shown to achieve state - of - the - art results on the SentEval tasks . \\ \midrule
        Draft & 
            Our proposal model can get both long - term dependence and local information well . \\
        \cmidrule(lr){1-1}  \cmidrule(lr){2-2}
        H-ND & 
            Our proposal can get both long - term and local information as well . \\
        \cmidrule(lr){1-1}  \cmidrule(lr){2-2}
        ED-ND &
            Our proposed model can capture both long - term dependencies and local information well . \\
        \cmidrule(lr){1-1}  \cmidrule(lr){2-2}
        GEC &
            Our proposal model can get both long - term dependence and local information well . \\
        \cmidrule(lr){1-1}  \cmidrule(lr){2-2}
        Reference & 
            Our proposed model can both capture long - term dependencies and local information well . \\

        \bottomrule
        \end{tabularx}
    }
\caption{Further examples of draft, reference, and the baseline models' output. }
\label{tbl:generation_2}
\end{table*}

\end{document}